\documentclass[conference, letterpaper, 10pt]{IEEEtran}
\usepackage{cite}
\usepackage{amsmath,amssymb,amsfonts}
\usepackage{algorithmic}
\usepackage{graphicx}
\usepackage{textcomp}
\usepackage{xcolor}
\def\BibTeX{{\rm B\kern-.05em{\sc i\kern-.025em b}\kern-.08em
    T\kern-.1667em\lower.7ex\hbox{E}\kern-.125emX}}
    
\usepackage{listings}
\definecolor{myblue}{HTML}{003399}
\definecolor{mygray}{HTML}{666666}
\lstdefinestyle{c++}{
    language = C++,
    basicstyle = \ttfamily \small,
    keywordstyle = \color{myblue},
    commentstyle = \itshape \color{mygray},
    morekeywords = {parfor, foreach, each, in, __shared__},
    escapeinside = {@}{@},
    showlines = true
}
\lstdefinestyle{caption}{
    style = c++,
    basicstyle = \ttfamily \footnotesize
}
\lstdefinestyle{listing}{
    style = c++,
    basicstyle = \linespread{1.3} \sffamily \small,
    literate=*{*}{\sffamily{*}}1,
    columns = fullflexible,
    keepspaces = true
}
\usepackage{tikz}
\usetikzlibrary{arrows, calc, matrix, positioning}
\usepackage{multirow}
\usepackage{hyperref}
\hypersetup{
    colorlinks=true,
    citecolor=black,
    linkcolor=black,
    urlcolor=myblue
}
\usepackage{tipa}

\begin{document}

\title{
Even Faster SNN Simulation with\\Lazy+Event-driven Plasticity and Shared Atomics
}


\def \CSD {CSD\,-\,UOC}
\def \ICS {FORTH\,-\,ICS}

\author{
\IEEEauthorblockN{1\textsuperscript{st} Dennis Bautembach}
\IEEEauthorblockA{\textit{\ICS{} \& \CSD{}}\\denniskb@ics.forth.gr}
\and
\IEEEauthorblockN{2\textsuperscript{nd} Iason Oikonomidis}
\IEEEauthorblockA{\textit{\ICS{}}\\oikonom@ics.forth.gr}
\and
\IEEEauthorblockN{3\textsuperscript{rd} Antonis Argyros}
\IEEEauthorblockA{\textit{\ICS{} \& \CSD{}}\\argyros@ics.forth.gr}}

\maketitle

\begin{abstract}
We present two novel optimizations that accelerate clock-based spiking neural network (SNN) simulators. The first one targets spike timing dependent plasticity (STDP). It combines lazy- with event-driven plasticity and efficiently facilitates the computation of pre- and post-synaptic spikes using bitfields and integer intrinsics. It offers higher bandwidth than event-driven plasticity alone and achieves a 1.5$\times$--2$\times$ speedup over our closest competitor. The second optimization targets spike delivery. We partition our graph representation in a way that bounds the number of neurons that need be updated at any given time which allows us to perform said update in shared memory instead of global memory. This is 2$\times$--2.5$\times$ faster than our closest competitor. Both optimizations represent the final evolutionary stages of years of iteration on STDP and spike delivery inside ``Spice'' (\textipa{/spaIk/}), our state of the art SNN simulator. The proposed optimizations are not exclusive to our graph representation or pipeline but are applicable to a multitude of simulator designs. We evaluate our performance on three well-established models and compare ourselves against three other state of the art simulators.
\end{abstract}


\section{Introduction}
SNNs are receiving a lot of attention since they more closely model biological neural networks than conventional ANNs do and therefore may prove more powerful. In contrast to ANNs, which can be inferred via matrix multiplications interleaved with activation functions, SNNs need to be \textit{simulated} because their neurons can undergo \textit{arbitrary} dynamics.

A SNN simulator works fairly simple:
\begin{enumerate}
    \item Update neurons, note which ones fire
    \item Update synapses
    \begin{itemize}
        \item Compute pre- and post-synaptic spikes
    \end{itemize}
    \item Deliver spikes to neurons' neighbors
\end{enumerate}

Neurons are exclusively governed by their dynamics and can thus be updated independently of one another, making it trivial to parallelize step (1) on GPUs. The synapse update requires more attention: In order to advance synapses' dynamics, a SNN simulator must first compute pre- and post-synaptic spikes which is a non-trivial task. Additionally, the sheer number of synapses (in the order of $O(\textit{\#neurons}^2)$) means that most of the simulation time will be spent in step~(2), requiring careful implementation that maps well to the hardware. Step~(3) is another non-trivial and expensive part of the simulation loop since again the number of spikes is in the order of $O(\textit{\#neurons}^2)$ and neurons are randomly interconnected, leading to poor memory access patterns.

In this paper we present two optimizations targeting steps~(2) and~(3) of the SNN simulation loop. In order to accelerate step~(2) we, for the first time, combine lazy plasticity~\cite{bautembach2020} with event-driven plasticity~\cite{Mattia2000,song2000competitive,morrison2007spike,Yavuz2016,Ahmad2018} and efficiently facilitate the computation of pre- and post-synaptic spikes via \mbox{per-neuron} bit fields and integer intrinsics. Combined, these measures yield a cumulative speedup of 30$\times$ over the na\"{i}ve approach of blindly updating each synapse at every loop iteration and are 1.5$\times$--2$\times$ faster than our closest competitor.

In order to accelerate step (3) we partition our adjacency list according to~\cite[Section III-D]{bautembach2021} into slices of equal width \textit{in the neuron domain}. Knowing that each slice indexes a fixed, contiguous subset of the neuron pool we are able to load said subset into shared memory and perform all updates there. The resulting spike delivery algorithm is 2$\times$--2.5$\times$ faster than relying on global memory and is \textit{at least} 2$\times$ faster than our closest competitor (depending on the benchmark).

These optimizations do not come at the cost of generality---users can still define custom models with arbitrary dynamics. Further, they are completely orthogonal to our multi-GPU acceleration scheme~\cite{bautembach2021}. Finally, they are not exclusive to our graph representation or pipeline but are applicable to a multitude of simulator designs.
\section{Related Work}

Tuckwell~\cite{tuckwell1988introduction} notes that early attempts to mathematically model and simulate the function of biological neurons are found since the beginning of the 20th century~\cite{McCulloch1943, lapique1907recherches}.
The research domain related with the efficient simulation of SNNs has a long history~\cite{Pelayo1997,Mattia2000,Reutimann2002} and is actively research to this day~\cite{VanDerVlag2019,Mozafari2019,bautembach2020,Panagiotou2020,Qu2020,Tian2021}.
Within the domain, important points of research focus include the improvement of biological fidelity~\cite{Chou2018} and numerical stability of methods~\cite{Tian2021}. Furthermore, hardware acceleration is explored~\cite{Smaragdos2017, Sripad2018}, including platforms such as VLSI~\cite{Pelayo1997}, FPGA~\cite{Sripad2018}, and even super-computers~\cite{Kunkel2014}.
On the theoretical front, advances include methods to quantify the difference between two spike trains~\cite{VanRossum2001}, exact solutions of differential equations that model membrane dynamics~\cite{Rudolph2006}, as well as a method to implement back propagation on SNNs~\cite{Tomlinson1990}.

\subsection{SNN Simulator Classification}
The various SNN simulators that have been proposed feature different key characteristics, strengths and weaknesses. Based on these traits, one can define useful classifications.
The biological fidelity level can be such a defining trait. According to it, simulators can be categorized into ones that aim to model the behavior of biological neurons as accurately as possible~\cite{Ros2006,Schemmel2010,Thibeault2011,Antonietti2016,Chou2018,Lee2018,VanDerVlag2019a}, and into those that are merely based on the general principle of spiking neurons.  

The versatility, and specifically the ability and ease of defining neuron models can serve as another way to categorize simulators. General purpose simulators allow the user to freely define the behavior of neurons and synapses~\cite{Goodman2010,Pecevski2014,Knight2018,Stimberg2018,Hazan2018,Yavuz2016,bautembach2020,bautembach2021}, whereas the rest only allow the use of a fixed number of predefined models~\cite{Schemmel2010,Ahmad2018,Chou2018,VanDerVlag2019a}.

Another important classification can be made based on the approach adopted regarding the high-level software implementation. In one category, the time-drive approach simulates the behavior of neurons and synapses in lock-step, advancing all of them in every iteration of the simulation~\cite{Ros2006a,Rudolph2006,Hanuschkin2010,Ahmad2018,Knight2018,bautembach2020,bautembach2021}. In contrast, event-driven simulators only update a neuron or synapse when a new event affects the (predictable, up until that time instant) evolution of their state~\cite{Ros2006a,Rudolph2006,Hanuschkin2010,Ahmad2018,Knight2018,bautembach2020}.
Ease of implementation and per-element efficiency are two strong points of the time-driven approaches, however if the time-step is set very low (desired to increase simulation accuracy), the simulation becomes computationally inefficient. Event-driven approaches on the other hand can achieve arbitrary time resolution with small or even negligible computational overhead, despite having low per-element computational efficiency.
As a middle ground, approaches that adopt characteristics of both categories can be called hybrid~\cite{Naveros2015,Naveros2018}. It is debatable whether approaches that adopt a time-driven neuron update with an event-driven synapse update (event-driven plasticity)~\cite{Mattia2000,song2000competitive,morrison2007spike,Yavuz2016,Ahmad2018} should be called hybrid, since this combination is less integral.

One last categorization can be made according to the target hardware platform of the simulator system. A wide variety of hardware has been employed, starting from regular CPUs~\cite{Reutimann2002,Delorme2003,Ros2006a,Plesser2007,Garrido2014,Pecevski2014,Panagiotou2020}, GPUs~\cite{bautembach2020,bautembach2021,Hazan2018,Qu2020,Mozafari2019,Stimberg2018,Sripad2018,Yavuz2016,Kasap2018,Ahmad2018,Szynkiewicz2016,Knight2018,VanDerVlag2019,VanDerVlag2019a,Thibeault2011}, combination of CPUs and GPUs~\cite{Krishnamani2010, Naveros2015,Naveros2018,Smaragdos2017}, and including even custom-built hardware~\cite{Ros2006,Pelayo1997, Schemmel2010, Furber2013, Kunkel2014, Sripad2018}.
Recent simulators are increasingly adopting the GPU architecture.

\subsection{Classification of Spice and Direct Competition}
According to the categorizations presented above, Spice~\cite{bautembach2020,bautembach2021} is general-purpose, time-driven, GPU-accelerated, and in this work it is fitted with event-driven plasticity.

For the quantitative evaluation of the proposed optimizations in our simulator, we choose as competitors three very recent works, closely related to ours: BSim~\cite{Qu2020}, GeNN~\cite{Yavuz2016}, and NeuronGPU~\cite{golosio2020new}. All three simulators are time-driven, GPU-accelerated, and, to varying degrees, general. This allows the implementation of the same SNN models on all of them, in turn enabling a direct comparison between them and our approach. We do not include event-driven simulators in the comparison since they pursue different goals with fundamentally different designs and imposed trade-offs, and so are not directly comparable. Even within the time-driven domain, the target of each simulator, and the associated performance can vary quite a lot. GeNN~\cite{Yavuz2016, Knight2018, Knight2021} has improved quite a bit since its introduction, and is usually among the fastest direct competitors. BSim~\cite{Qu2020} among other features supports multi-GPU setups. Finally, NeuroGPU~\cite{golosio2020new} is the least comparable of the three simulators. It strives to maximize biological fidelity and as such features double precision arithmetic and ``exact integration''~\cite{Rotter1999}. We still included it in this comparison, mostly for completeness, to gauge what performance trade-off these design choices entail.
\begin{figure}
    \centering
    \tikzstyle{node} = [anchor = center, draw, minimum size = 1.5em, very thin]
    \tikzstyle{arr} = [-{stealth'}, thin]
    \tikzstyle{split} = [thick, red!80]
    \tikzstyle{split2} = [thick, dotted, red!90]
    \begin{tikzpicture}[node distance = 1.2cm and 1cm, x = 1.5em, y = 1.5em]
        \matrix (adjhead) [
            matrix of nodes,
            nodes = node
        ]{    
            0 & 1 & 2 & 4 & 6 & 7 \\
            0 & 1 & 4 & 5 & 7 & 8 \\
        };
        
         \matrix (adjtail) [
            below = of adjhead,
            matrix of nodes,
            nodes = node
        ]{    
            1 & 3 & 4 & 5 & 6 & $\infty$ \\
        };
        
        \matrix (pivhead) [
            right = of adjhead,
            matrix of nodes,
            nodes = {node, text=red!90}
        ]{    
            0 & 3 & 4 & 6 \\
            0 & 2 & 4 & 6 \\
        };
        
         \matrix (pivtail) [
            below = of pivhead,
            matrix of nodes,
            nodes = {node, text=red!90}
        ]{    
            0 & 1 & 4 & 5 \\
        };
        
        \matrix (pool) [
            below = of adjtail.center,
            matrix of nodes,
            nodes = node
        ]{
            n & e & u & r & o & n & a & r & r\\ 
        };
        
        \node [anchor=south] at (adjhead.north) {Adjacency List};
        \node [anchor=south] at (pivhead.north) {Pivots};
        \node (Syn) [anchor=north east, text width=1cm, align=center] at (adjhead.south west) {Synapse Array};
        \node (N) [anchor=north west] at ($ (pivhead.south east) + (.5,0) $) {$N$};
        
        \foreach[count=\src] \dst in {2,4,5,6,7}
            \draw [arr] (adjtail-1-\src.south) to [out=270,in=90] (pool-1-\dst.north);
            
        \draw [split] (adjhead-1-1.north west) -- (adjhead-1-1.south west);
        \draw [split] (adjhead-1-4.north west) -- (adjhead-1-4.south west);
        \draw [split] (adjhead-1-5.north west) -- (adjhead-1-5.south west);
        \draw [split] (adjhead-1-6.north east) -- (adjhead-1-6.south east);
        \draw [split, very thick, dotted] (adjhead-1-3.south west) -- (adjhead-1-3.south east);
        
        \draw [split] (adjhead-2-1.north west) -- (adjhead-2-1.south west);
        \draw [split] (adjhead-2-3.north west) -- (adjhead-2-3.south west);
        \draw [split] (adjhead-2-5.north west) -- (adjhead-2-5.south west);
        \draw [split] (adjhead-2-6.north east) -- (adjhead-2-6.south east);
        
        \draw [split] (adjtail-1-1.north west) -- (adjtail-1-1.south west);
        \draw [split] (adjtail-1-2.north west) -- (adjtail-1-2.south west);
        \draw [split] (adjtail-1-5.north west) -- (adjtail-1-5.south west);
        \draw [split] (adjtail-1-6.north west) -- (adjtail-1-6.south west);
        
        \draw [split] (pool-1-3.north east) -- (pool-1-3.south east);
        \draw [split] (pool-1-6.north east) -- (pool-1-6.south east);
        
        \draw [very thin] ($ (adjhead-2-1.west) + (0,-.25) $) -- +(-.25,0) -- +(-.25,2) -- +(5.75,2) -- +(5.75,1.75);
        \draw [very thin] ($ (adjtail-1-1.west) + (0,-.25) $) -- +(-.25,0) -- +(-.25,1) -- +(5.75,1) -- +(5.75,.75); 
        
        \path (adjhead) -- node[auto=false]{\vdots} (adjtail);
        \path (pivhead) -- node[auto=false]{\vdots} (pivtail);
        
        \draw[thin] (Syn.north) -- (adjhead.west);
        \draw[thin] (Syn.south) -- (adjtail.north west);
        
        \draw[thin] (pivhead.north east) to [out=315,in=135] (N.west);
        \draw[thin] (pivtail.south east) to [out=45,in=225] (N.west);
        
        \draw[split2] (adjhead-2-1.south west) -- (adjtail-1-1.north west);
        \draw[split2] (adjhead-2-3.south west) -- (adjtail-1-2.north west);
        \draw[split2] (adjhead-2-5.south west) -- (adjtail-1-5.north west);
        \draw[split2] (adjhead-2-6.south east) -- (adjtail-1-6.north west);
        
        \draw[split2] (adjtail-1-1.south west) -- (pool-1-1.north west);
        \draw[split2] (adjtail-1-2.south west) -- (pool-1-4.north west);
        \draw[split2] (adjtail-1-5.south west) -- (pool-1-7.north west);
        \draw[split2] (adjtail-1-6.south west) -- (pool-1-9.north east);
    \end{tikzpicture}
    \caption{Overview over our data structures. We represent the graph using a padded/rectangular adjacency list. It has $N=|\textit{neurons}|$ many rows. Row $i$ represents neuron $i$'s outgoing connections in the form of indices into the neuron array. They are sorted. Non-connections are represented by a special value (\lstinline[style=caption]{INT_MAX} in practice). The synapse array has the same number of entries as the adjacency list with an implicit 1:1 mapping between the two (\lstinline[style=caption]{adj[i][j]} corresponds to \lstinline[style=caption]{synapses[i][j]}). We divide the neuron array into chunks of even size, in this case 3. We then binary-search pivots so that entries \lstinline[style=caption]{adj[i][pivots[i][j]..pivots[i][j+1]-1]} only index neurons $[3j, 3j+3)$. We explicitly store the implicit ``0'' so as to avoid special cases in the implementation.}
    \label{figure:adj}
\end{figure}
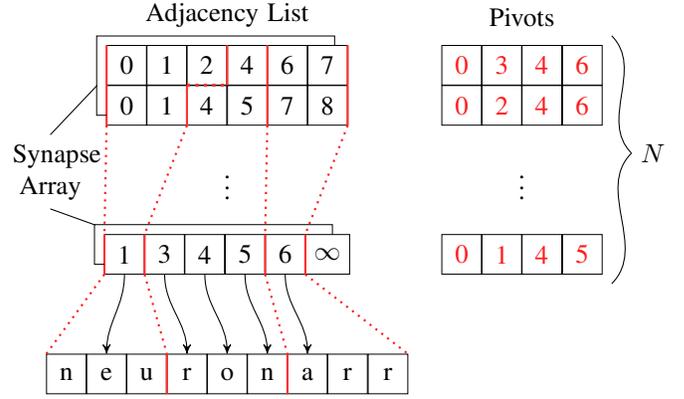

\section{Method}
SNNs are essentially directed graphs (neurons = vertices, synapses = edges). They are typically represented as a (binary) adjacency matrix, adjacency list, or compressed sparse row (CSR). All representations are grouped by source neuron (that is, a neuron's outgoing connections are stored in contiguous memory) because it is suitable for computations. The connections are also sorted to improve cache coherency. Since a SNN's topology is static, the graph needs to be constructed only once at the beginning of a simulation.

We will illustrate the optimization techniques described in this paper using a (padded/rectangular) adjacency list because it is a very simple representation and also happens to be what Spice uses (Fig.~\ref{figure:adj}). The presented optimizations are applicable to a variety of representations though, including all of the commonly used ones mentioned above.

\begin{figure*}
\begin{lstlisting}[xleftmargin=2em,style=listing]
extern int delay; // synaptic delay (in time steps) of the network
struct neuron { int64 hist; ... }; // firing history, each bit represents one time step, LSB = most recent
struct synapse { neuron& src; neuron& dst; ... };
void update(synapse& syn, bool preSynapticSpike, bool postSynapticSpike, int numSteps = 1); // user-defined callback
\end{lstlisting}
\begin{minipage}{.36\textwidth}
\begin{lstlisting}[numbers=left,xleftmargin=2em,style=listing]
parfor n in neurons:
  parfor syn in outgoingSynapses(n):
  
  
  
    update(
      syn,
      n.hist[delay],
      syn.dst.hist[0]);
      
      
      
\end{lstlisting}
\centering
(a)
\end{minipage}
\begin{minipage}{.33\textwidth}
\begin{lstlisting}[style=listing]
parfor n in @\textbf{firing}@ neurons:
  parfor syn in outgoingSynapses(n):
    age = now - n.timeOfLastUpdate;
    
    for i in age..0:
      update(
        syn,
        i == 0,
        syn.dst.hist[i]);
        
        
        
\end{lstlisting}
\centering
(b)
\end{minipage}
\begin{minipage}{.3\textwidth}
\begin{lstlisting}[style=listing]
parfor n in firing neurons:
  parfor syn in outgoingSynapses(n):
    age  = now - n.timeOfLastUpdate;
    j = -1;
    for i in setBits(syn.dst.hist[age..0]):
      update(
        syn,
        i == age,
        true,
        @\textbf{i - j}@);
      j = i;
    update(syn, j < age, false, age - j);
\end{lstlisting}
\centering
(c)
\end{minipage}\\
\caption{Pseudocode showing the evolution of our implementation from na\"{i}ve, over lazy, to event-driven plasticity. \textbf{(a)  Na\"{i}ve plasticity}. We loop over all synapses (lines 1--2) and invoke the user-provided callback \lstinline[style=caption]{update()} (line 6) which is part of the model definition. We pass it a reference to each synapse (line 7) as well as information about the occurrence of pre- and post-synaptic spikes by reading the corresponding bits from the source and destination neuron's firing histories (lines 8--9, \lstinline[style=caption,mathescape]!x[i] := (x $\gg$ i) & 1!). The \lstinline[style=caption]{parfor} loops can be trivially parallelized because each loop iteration is independent. In the case of two nested \lstinline[style=caption]{parfor}s the outer loop iterations could be assigned to different CUDA blocks and the inner loop iterations could be assigned to different threads within a block. \textbf{(b) Lazy plasticity}. Instead of updating all synapses, we only update those originating from firing neurons (line 1) since they are about to transmit a spike and thus need be up to date. We compute the synapses' ``age'', i.e. the number of iterations for which we have intentionally neglected them (line 3) and "replay" all the missed updates (line 5). \textbf{(c)  Event-driven plasticity} by contrast does not replay the entire firing history since the last spike, but only iterates over set bits, i.e. post-synaptic spikes, skipping ahead multiple update steps in-between (line 5). This is communicated to the user via \lstinline[style=caption]{update()}'s fourth parameter which represents the number of time steps that have elapsed since the callback was last invoked (line 10). The \lstinline[style=caption]{update()} call in line 12 handles the "tail".
\vspace{.1cm}
}

\label{figure:pseudo_plastic}
\end{figure*}

\subsection{Lazy+Event-driven Plasticity}
\label{section:plasticity}
``Plasticity'' refers to models with changing (i.e. plastic) synapse state---neurons are always plastic. The sheer number of synapses makes this a very costly operation worth optimizing. The na\"{i}ve approach (Fig.~\ref{figure:pseudo_plastic}(a)) would be to simply update every synapse at each simulation step. This is incredibly slow with the limiting factor being memory bandwidth: Brunel+ uses 12~B per plastic synapse which make up 40\% of all synapses. In order to run a network with 1B synapses total in real time, it would require 96~TB/s of bandwidth (every plastic synapse would have to be read and written 10K times per second of biological time). In comparison, a NVIDIA Tesla V100 offers $\sim$0.7~TB/s of bandwidth.

We already presented a simple yet effective optimization to this na\"{i}ve approach called ``Lazy Plasticity''~\cite[Section III-A1]{bautembach2020} (Fig.~\ref{figure:pseudo_plastic}(b)). It takes advantage of the facts that
\begin{itemize}
    \item neurons fire infrequently and
    \item a synapse need only be up to date to transmit a spike
\end{itemize}
Rather than updating all synapses at every step we intentionally keep them in a stale state. When a neuron fires, we compute its $\textit{age}=\textit{now}-\textit{timeOfLastSpike}$ and perform that many update steps on its outgoing synapses. While the total number of computations stays the same, they can now be performed inside registers with only the initial and final state having to be read from/written to global memory, increasing our effective bandwidth (Section~\ref{section:result_opt}).

Plasticity need not only update synapses' states but also facilitate the efficient computation of pre- and post-synaptic spikes. A pre-synaptic spike occurs if a synapse's source neuron fires, a post-synaptic spike occurs if a synapse's destination neuron fires. We must inform the synapse of both types of events. Pre-synaptic spikes are trivial to compute, especially if synapses are grouped by source neuron: The fact that a neuron fires, implies that all its outgoing synapses experience a pre-synaptic spike (potentially after some delay). Luckily for us these same synapses are already laid out in a known, contiguous region of memory, making it very efficient to iterate over and update them.

Post-synaptic spikes are more tricky to compute since we need to consolidate a neuron's \textit{incoming} synapses. An infeasibly slow solution would be to iterate over the \textit{entire} adjacency list and search for all occurrences of said neuron, and repeat this for all firing neurons. Some simulators~\cite{Ahmad2018,Yavuz2016} tackle this issue by storing a ``reverse'' adjacency list. That is, in addition to storing outgoing synapses they also store incoming synapses per neuron. This has several disadvantages:
\begin{itemize}
    \item It doubles the memory footprint of the graph representation.
    \item While the adjacency data of incoming synapses are now laid out in contiguous memory and can thus be efficiently iterated, the synapses they index are spread all over memory resulting in very poor bandwidth when updating them.
\end{itemize}

We instead store the network's recent firing history in the form of \mbox{per-neuron} bit fields (typically 64 bit large). When it is time to update a row of synapses (because their source neuron fired), we read each destination neuron's firing history and loop over the most recent \textit{age} bits which represent the post-synaptic spikes. This is also sub-optimal in terms of bandwidth since the read accesses are scattered over memory, however,
\begin{itemize}
    \item they are confined to a smaller region of memory (kilobytes vs.\ gigabytes) resulting in better cache utilization,
    \item their addresses are steadily increasing due to the sorted nature of the adjacency list, resulting in better cache coherency, and
    \item they are only performed once for (up to) 64 update steps, increasing our effective bandwidth.
\end{itemize}

From here, implementing event-driven plasticity is trivial: Instead of looping over all \textit{age} bits, we only loop over the set ones which can be done very efficiently using integer intrinsic \lstinline{__clz()} (Fig.~\ref{figure:pseudo_plastic}(c)). In order to benefit from event-driven plasticity, synapse dynamics must have a closed-form solution so they can "skip ahead" multiple steps at a time. If this is not possible, users of our simulator can simply loop internally in which case we gracefully degrade back to lazy plasticity.

\subsection{Spike Delivery with Shared Atomics}
\label{section:spike_delivery}
Whenever a neuron fires, its neighbors must be informed (potentially after a model-specific delay). The na\"{i}ve approach of (in parallel, using a CUDA block) looping over the corresponding row in the adjacency list and invoking the user-defined \lstinline{deliver()} callback for each neighbor (Fig.~\ref{figure:pseudo_static}(a)), works fairly well: Reads from the adjacency list get fully coalesced. Writes to the neurons \textit{are} scattered, the negative effects of which get mostly mitigated by the cache. 
\footnote{Writes are scattered so far apart that we would expect our effective bandwidth to be $^1/_{32}$nd of the theoretical bandwidth as each 4~B word read/write would result in its own memory transaction (128~B large on CUDA devices). Thanks to the cache, this is far from the case in practice.}

In~\cite[Section III-C]{bautembach2021} we introduced an effective optimization which we call ``cache-aware'' spike delivery: Instead of traversing the adjacency list row-wise, we traverse it per column or, more accurately, per \mbox{32-column} slice so as to maintain full coalescing of the adjacency data. While writes are still scattered, they are now statistically expected to be closer together, resulting in a better cache hit rate. Cache-aware spike delivery is strictly faster than the na\"{i}ve one but only really shines once neurons stop fitting into cache. At that point na\"{i}ve delivery starts to grow quadratically with network size while cache-aware delivery continues to grow linearly. Cache-aware spike delivery is slightly sub-optimal in the sense that metadata (such as information about the source neuron) need to be re-read for every warp whereas previously they only had to be read once per block. This is negligible in practice though.

\begin{figure}
\begin{lstlisting}[style=listing]
// adjacency list
extern int adj[/*rows*/][/*columns*/];
// offsets into 'adj' marking the start index of each slice
extern int pivots[/*rows*/][/*columns*/];
struct neuron { int id; ... };
// user-defined callback
void deliver(const neuron& source, neuron& destination);
\end{lstlisting}

\begin{minipage}{0.94\columnwidth}
\begin{lstlisting}[numbers=left,xleftmargin=2em,style=listing]
parfor src in firing neurons:
    parfor dst in neighbors(src):
        deliver(src, dst);
\end{lstlisting}
\end{minipage}(a)

\begin{minipage}{0.94\columnwidth}
\begin{lstlisting}[numbers=left,xleftmargin=2em,style=listing]
parfor slice in 0..|neurons| / 1024 - 1:
    __shared__ neuron shared[1024];

    parfor i in 0..1023:
        shared[i] = neurons[slice * 1024 + i];

    parfor src in firing neurons:
        parfor dstID in adj[src.id][
            pivots[src.id][slice]..
            pivots[src.id][slice+1]-1]:
            deliver(src, shared[dstID % 1024]);

    parfor i in 0..1023:
        neurons[slice * 1024 + i] = shared[i];
\end{lstlisting}
\end{minipage}(b)

\caption{Pseudocode contrasting na\"{i}ve with shared memory-based spike delivery. \textbf{(a) Na\"{i}ve delivery}. We loop over all firing neurons (line 1) and inform their neighbors (line 2) about the spike by invoking the user-provided callback \lstinline[style=caption]{deliver()} (line 3) which is part of the model definition. \textbf{(b) Shared memory-based delivery}. We cooperatively load a 1024-neuron slice from global into shared memory (lines 4--5). Once again we deliver all spikes, but only to neurons that fall into our slice. The delivery is carried out in shared memory (lines 7--11). When done we write the new neuron state back to global memory (lines 13--14). In practice we assign a 1024-thread block to each slice, establishing a 1:1 mapping between threads and neurons which greatly simplifies code (the \lstinline[style=caption]{parfor}s in lines 4 \& 13 turn into simple assignments). We then assign each warp within the block to one spike and finally each thread within the warp to one destination neuron. This code assumes that the total number of neurons is a multiple of 1024, for brevity. Additional logic is needed to guard against out-of-bounds accesses.}
\label{figure:pseudo_static}
\end{figure}

\def\width{0.67\columnwidth}
\pgfmathsetlengthmacro\height{0.8*\width}
\pgfmathsetlengthmacro\gap{\columnwidth-\width}
\begin{figure*}
    \centering
    \includegraphics[width=\width,height=\height]{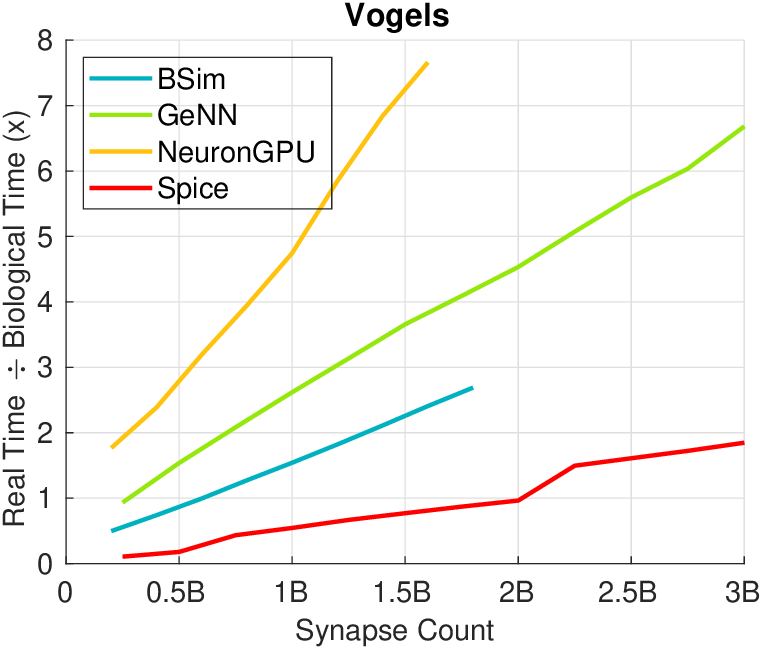}
    \includegraphics[width=\width,height=\height,trim=6mm 0 0 0,clip]{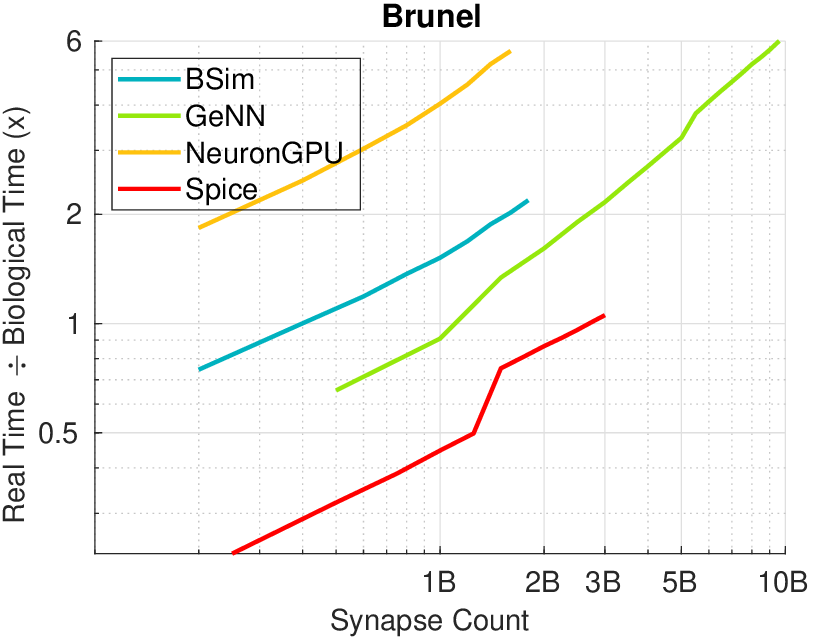}
    \includegraphics[width=\width,height=\height,trim=6mm 0 0 0,clip]{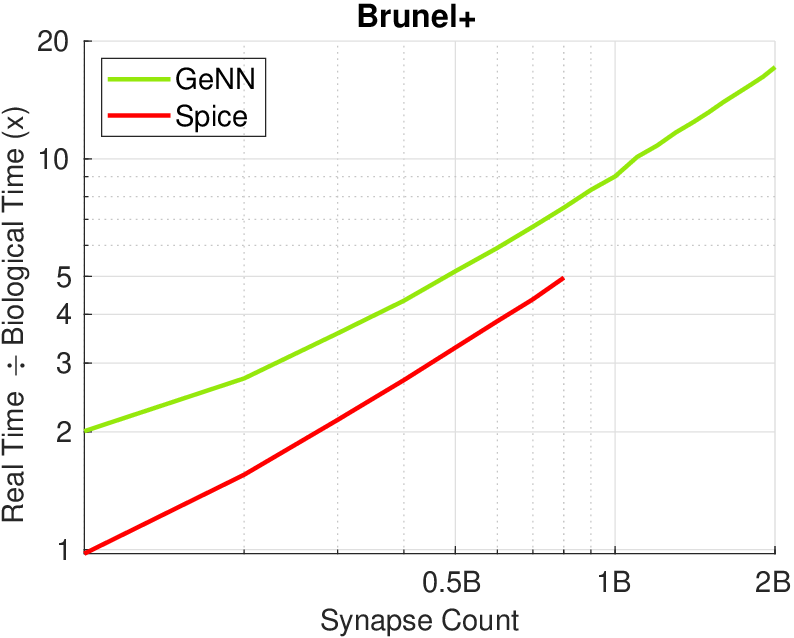}
    \caption{Simulation time as a function of network size for each model: We measure the time it takes to simulate 10~s of biological time for various synapse counts and report $\textit{wall-clock time} \div \textit{biological time}$. Note that in the Brunel(+) case both axes are logarithmic.}
    \label{figure:simtime}
\end{figure*}

In this paper we drive cache-aware spike delivery to its conclusion: We logically divide the neuron domain into chunks of equal size, say 1024 neurons each. We then partition the adjacency list into slices that index a single and only a single chunk, which is simply achieved by pre-computing $\lceil \frac{n}{1024} \rceil$ pivots per row via binary search (Fig.~\ref{figure:adj}(red dotted lines)). This turns our statistical expectation into a mathematical certainty: Knowing that slice $i$ only indexes neurons $[i*1024, (i+1)*1024)$ allows us to load them into shared memory and deliver all spikes there (Fig.~\ref{figure:pseudo_static}(b)). Doing so:
\begin{itemize}
    \item minimizes global memory traffic (only the initial/final neuron states have to be read from/written to global memory),
    \item uses faster shared memory,
    \item uses faster shared atomics, and
    \item circumvents the cache entirely (avoiding cache misses/evictions/etc.)
\end{itemize}
Shared memory-based spike delivery is strictly faster than cache-aware spike delivery (regardless of the number of neurons) for practical network densities and shared memory sizes.
\begin{figure*}
    \centering
    \includegraphics[width=\width,height=\height]{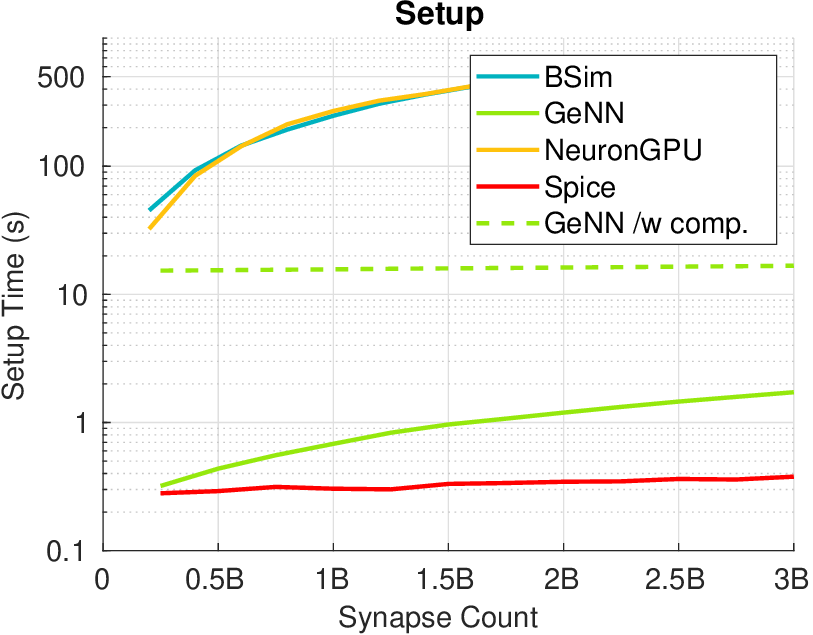}
    \includegraphics[width=\width]{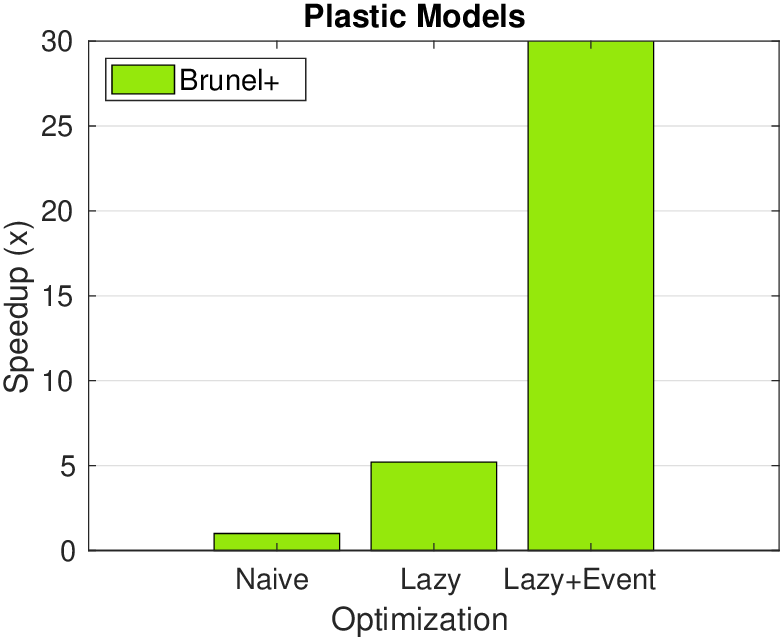}
    \includegraphics[width=\width]{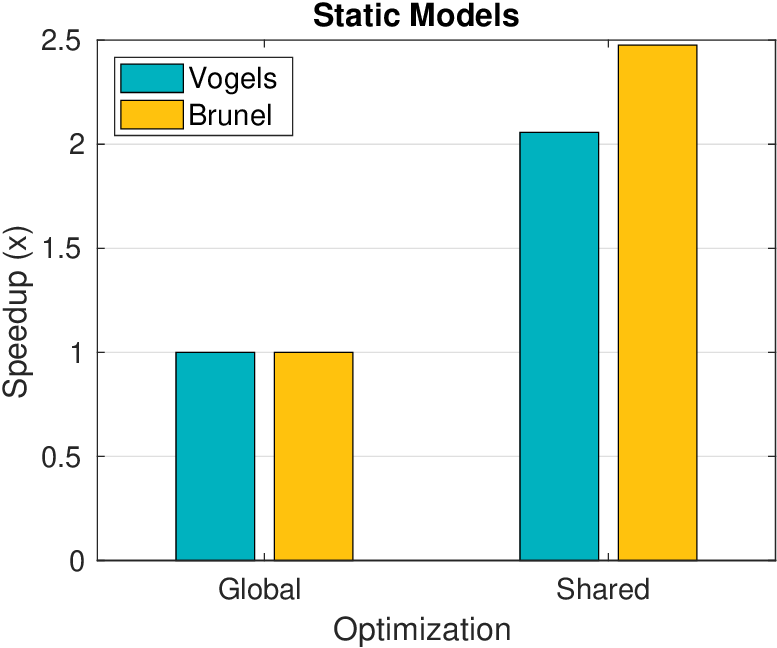}
    \caption{(left) Setup time as a function of network size for Vogels. The y-axis is logarithmic. The green, dashed line represents setup time + compilation time.
    (middle) Speedup due to various plasticity optimizations. The baseline is na\"{i}ve plasticity which updates every synapse at each simulation step. Shown are the speedups achieved due to lazy plasticity alone, as well as lazy plasticity combined with event-driven plasticity. (right) Speedup due to shared memory-based spike delivery. The baseline is na\"{i}ve spike delivery relying on global atomics and the cache.}
    \label{figure:setup_speedup}
\end{figure*}

\section{Results}
\begin{figure}[h]
    \centering
    \includegraphics[width=0.45\columnwidth,trim=40mm 10mm 35mm 10mm,clip]{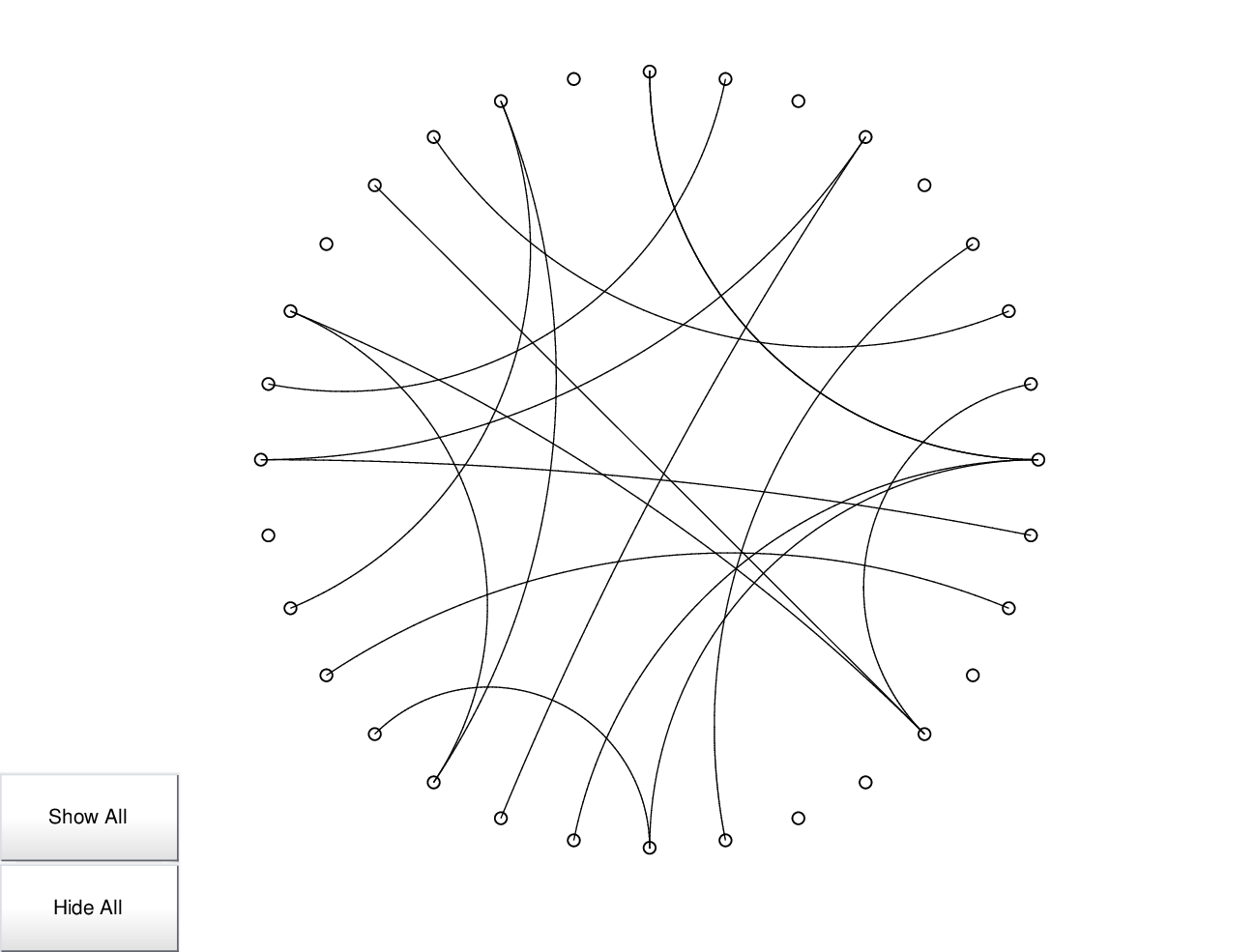}
    \hspace{0.07\columnwidth}
    \includegraphics[width=0.45\columnwidth,trim=40mm 10mm 35mm 10mm,clip]{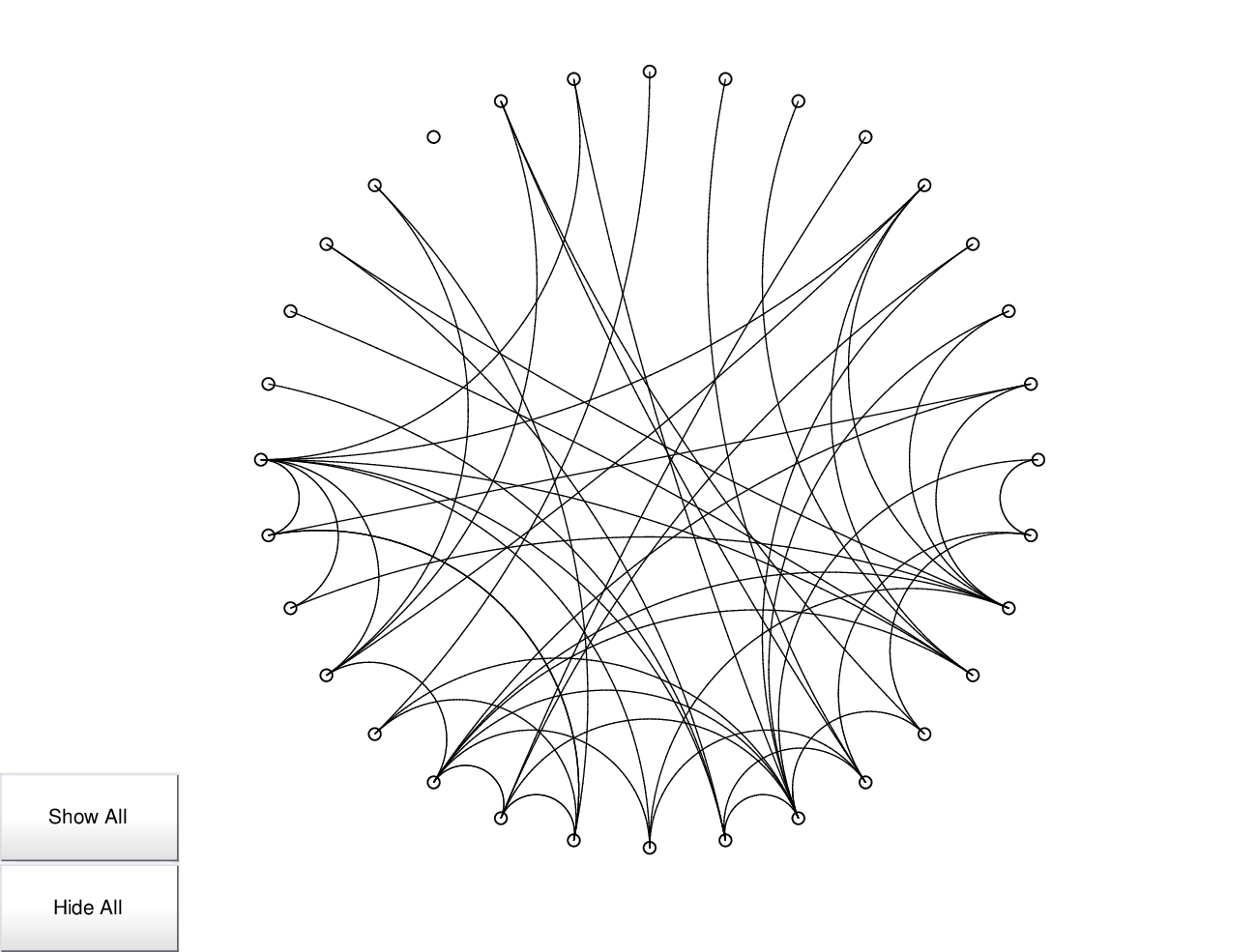}
    \includegraphics[width=\columnwidth,trim=23mm 15mm 25mm 5mm,clip]{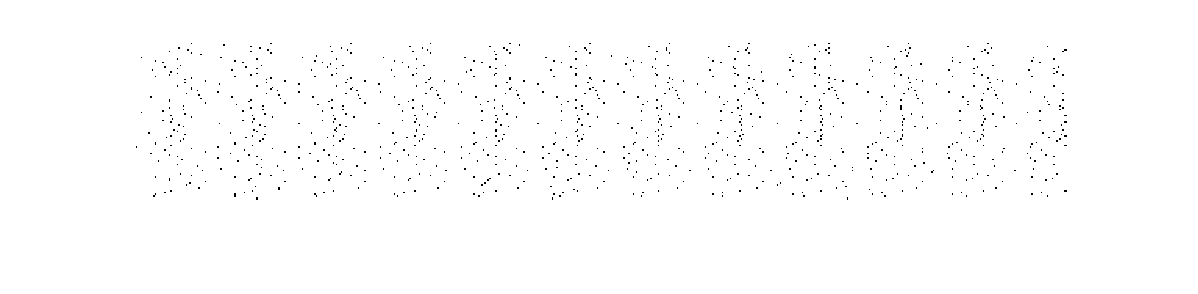}\\
    \vspace{2mm}
    \includegraphics[width=\columnwidth,trim=23mm 15mm 25mm 5mm,clip]{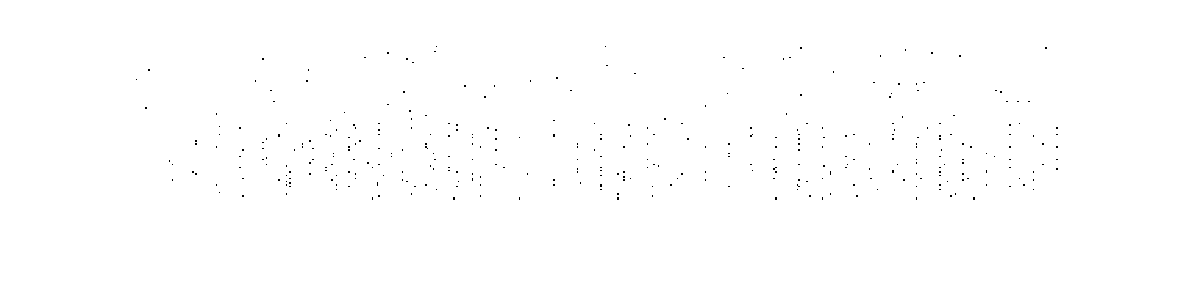}
    \caption{Visualization of Vogels and Brunel(+). (top left) Graph of Vogels. (top right) Graph of Brunel(+). (middle) Firing pattern of Vogels. Each row represents a neuron, each column a time step, each dot a spike. (bottom) Firing pattern of Brunel (Brunel+ fires very similarly).}
    \label{fig:modelviz}
\end{figure}

We compare the performance of our simulator to that of BSim~\cite{Qu2020}, GeNN~\cite{Yavuz2016}, and NeuronGPU~\cite{golosio2020new} using three well-established models: Vogels-Abbott (``Vogels'') and Brunel (with and without plasticity, ``Brunel(+)''), detailed in~\cite[Appendixes A \& B]{Ahmad2018}. We apply a scaling factor to synaptic weights allowing us to vary the network size while maintaining the overall firing pattern~\cite{bautembach2020}. The models are visualized in Fig.~\ref{fig:modelviz}.

When comparing ourselves to other simulators we enable all optimizations. The impact of individual optimizations is analyzed in Section~\ref{section:result_opt}.

All the code used for the experiments can be found at:
\begin{itemize}
    \item \textbf{BSim} \href{https://github.com/denniskb/bsim}{github.com/denniskb/bsim}, forked from master as of Feb 19, 2020.
    \item \textbf{GeNN} \href{https://github.com/denniskb/genn}{github.com/denniskb/genn}, forked from tag ``GeNN 4.4.0'' as of Jan 5, 2021.
    \item \textbf{NeuronGPU} \href{https://github.com/denniskb/neurongpu}{github.com/denniskb/neurongpu}, forked from master as of Oct 20, 2020.
    \item \textbf{Spice} \href{https://github.com/denniskb/spice/tree/gather}{github.com/denniskb/spice/tree/gather}, as of May 7, 2021.
\end{itemize}

All benchmarks were performed on a Google Cloud VM with an Intel Xeon E5-2699 v3, a Nvidia Tesla V100 16 GB, and 256 GB RAM, running a headless Ubuntu 20 with CUDA 11 and GCC 9.

\subsection{Simulation Time as a Function of Network Size}
We measure the time it takes to simulate 10~s of biological time for various network sizes (synapse counts). We report $\textit{wall-clock time}\div\textit{biological time}$ (Fig.~\ref{figure:simtime}).

NeuronGPU uses double precision arithmetic and ``exact integration''~\cite{Rotter1999} as opposed to single precision arithmetic and Euler integration used by the other simulators---we do not expect it to perform on par but include it for completeness, to gauge how much performance one has to sacrifice for biological fidelity.

GeNN offers the choice between three different connectivity types: \lstinline{SPARSE_GLOBALG} is similar to an adjacency list, \lstinline{BITMASK_GLOBALG} is similar to a binary adjacency matrix, and \lstinline{PROCEDURAL_GLOBALG} does not store the graph at all but generates it on the fly. We found that \lstinline{SPARSE_GLOBALG} is faster for Vogels while \lstinline{BITMASK_GLOBALG} is faster for Brunel(+). The latter is also more memory efficient for dense networks, allowing GeNN to simulate much larger instances of Brunel(+).

While both BSim and NeuronGPU support STDP, their synapse types do not quite match the behavior of Brunel+. According to the authors, modifying them ``currently is a task for developers, not for users''.

\subsection{Setup Time as a Function of Network Size}
Fast setup is important as it allows more experiments to run in quick succession and thus speeds up network design and parameter tuning. We measure the time in seconds it takes to initialize Vogels for various network sizes (Fig.~\ref{figure:setup_speedup}(left)). GeNN and Spice perform this on the GPU which is why they are orders of magnitude faster. In GeNN, \textit{any} parameter change requires recompilation which takes $\sim$15~s on the test machine (green, dashed line). Our network construction is so fast ($\sim$200M synapses/\textbf{ms}) that setup is dominated by memory allocations and thus virtually constant w.r.t. network size.

BSim and NeuronGPU also use a lot of RAM during setup, peaking at 200 GB for BSim and 40 GB for NeuronGPU~\cite[Table 1]{bautembach2021}, which might be prohibitive for some users.

\subsection{Impact of Optimizations on Simulation Time}
\label{section:result_opt}
In the preceding plots all optimizations were enabled for Spice. In this section we analyze the relative speedup that each optimization contributes to the final simulation time, compared to previous implementations (Fig.~\ref{figure:setup_speedup}(middle+right)).

To the best of our knowledge, nobody uses na\"{i}ve plasticity due to how slow it is (Fig.~\ref{figure:setup_speedup}(middle)). However, it makes for a good baseline because it is very close to a ``first implementation'' when porting/writing an algorithm to/in CUDA. Lazy plasticity (with a firing history of 64 steps) is 5 times faster. Note that both algorithms perform the exact same number of computations. The speedup is only due to decreased global memory traffic. Beyond 64 steps, diminishing returns kick in: While the effective bandwidth continues to increase, dynamics' computational costs stay constant and start to dominate the simulation time. Going from 64 to 128 steps is only $\sim$15\% faster while it increases code verbosity (emulating \mbox{128-bit} integers) and might not even generalize to other, higher-activity models. Event-driven plasticity is another 6 times faster, which might seem little when considering that most neurons reach their maximum age and their firing histories only have 0--1 bits set. However, internal tests revealed that we are operating close to the theoretical limit of our approach: When bypassing the event-driven plasticity logic entirely and "blindly" updating synapses in a single step, we only observed a 25\% increase in performance. That is not to say that there is not a \textit{fundamentally different}, potentially much faster approach.

Once again, the straightforward, row-wise spike-delivery algorithm serves as the baseline (Fig.~\ref{figure:setup_speedup}(right)). Shared memory-based spike delivery achieves its 2$\times$--2.5$\times$ speedup with a slice width of 1024 neurons. The slice width is a delicate, tunable parameter. Increasing it increases the average gap between pivots which leads to better memory bandwidth when reading adjacency data and higher parallelism when delivering spikes. At the same time though, it leads to higher shared memory consumption which can inhibit parallelism again. 1024 neurons strike a good balance in our benchmarks and have the additional benefit of allowing us to map them 1:1 to CUDA threads, simplifying code.
\section{Summary and Future Work}
We presented two algorithms for the efficient facilitation of STDP and spike delivery inside SNNs, which significantly outperform the state of the art. They can be retrofitted to existing simulators with minimal code additions and without the need for fundamental changes to the code architecture. STDP requires the allocation and update of per-neuron firing histories and in turn makes reverse adjacency lists obsolete. Spike delivery requires the computation of pivots which index the existing graph representation. In the case of adjacency matrices, the pivots do not even have to be stored but can be inferred.

A great feature that would interoperate seamlessly with our presented optimizations is graph compression. The way we partition the adjacency list lends itself to a very simple block compression scheme: Since every entry indexes only 1024 neurons, 10 bits are sufficient to represent it. If we packed 3 consecutive indices into one \mbox{32-bit} integer, we could reduce memory consumption by almost $^2/_3$rds. This would allow us to triple the size of static models regardless of their density.

\bibliographystyle{IEEEtran}
\bibliography{manual, mendeley}

\end{document}